\documentclass[journal]{IEEEtran}

\usepackage[english]{babel}

\usepackage{soul}

\usepackage{tablefootnote}
\usepackage{acronym}
\usepackage{hyperref}
\usepackage{booktabs}
\usepackage[normalem]{ulem}
\useunder{\uline}{\ul}{}
\usepackage{amsmath}
\usepackage{amsfonts} 
\usepackage{amssymb}
\usepackage{mathtools}
\usepackage{pdfpages}
\usepackage{booktabs}
\usepackage{subcaption}
\usepackage{xurl} 
\usepackage{xstring} 
\usepackage{multirow}

\hypersetup{
    colorlinks=true,
    linkcolor=red,
    anchorcolor=black,
    citecolor=green,
    filecolor=magenta,      
    urlcolor=cyan,
    pdftitle={},
    pdfauthor={Anonymous},
    }

\input{sty/macros.sty}
\input{sty/acronyms.sty}

\renewcommand{\figref}[1]{Figure~\ref{#1}} 	
\renewcommand{\figref}[1]{Figure~\ref{#1}}

\input{sty/acronyms.sty}
\begin{document}
    %
\title{Data-driven Super-Resolution of Flood Inundation Maps using Synthetic Simulations}
%
%
%

\author{Akshay~Aravamudan,
        Zimeena~Rasheed,
        Xi~Zhang,
        Kira~E.~Scarpignato,\\
        Efthymios~I.~Nikolopoulos,
        Witold~F.~Krajewski,
        and Georgios~C.~Anagnostopoulos \IEEEmembership{Senior Member, IEEE}
        
\thanks{A. Aravamudan, X. Zhang, K. E. Scarpignato and G. C. Anagnostopoulos are with the Computer and Engineering Sciences Department at Florida Institute of Technology\\
Z. Rasheed and E. I. Nikolopoulos are with the Civil and Environmental Engineering at Rutgers University\\
W. F. Krajewski is with the Civil and Environmental Engineering at University of Iowa}
\thanks{Manuscript received XX XX, XXXX; revised XX XX, 20XX.}}

%
%

\markboth{Journal of \LaTeX\ Class Files,~Vol.~14, No.~8, August~2015}%
{Shell \MakeLowercase{\textit{et al.}}: Bare Demo of IEEEtran.cls for IEEE Journals}

    \maketitle
    \begin{abstract}
The frequency of extreme flood events is increasing throughout the world. Daily, high-resolution (30m) \acp{FIM} observed from space play a key role in informing mitigation and preparedness efforts to counter these extreme events. However, the temporal frequency of publicly available high-resolution \acp{FIM}, \textit{e.g.}, from Landsat, is at the order of two weeks thus limiting the effective monitoring of flood inundation dynamics. Conversely, global, low-resolution ($\sim$300m) \acp{WFM} are publicly available from NOAA VIIRS daily. Motivated by the recent successes of deep learning methods for single image super-resolution, we explore the effectiveness and limitations of similar data-driven approaches to downscaling low-resolution \acp{WFM} to high-resolution \acp{FIM}. To overcome the scarcity of high-resolution \acp{FIM}, we train our models with high-quality synthetic data obtained through physics-based simulations. We evaluate our models on real-world data from flood events in the state of Iowa. The study indicates that data-driven approaches exhibit superior reconstruction accuracy over non-data-driven alternatives and that the use of synthetic data is a viable proxy for training purposes. Additionally, we show that our trained models can exhibit superior zero-shot performance when transferred to regions with hydroclimatological similarity to the U.S. Midwest. 
\end{abstract}

\begin{IEEEkeywords}
flood inundation maps, super-resolution, water fraction maps, machine learning
\end{IEEEkeywords}

    \IEEEpeerreviewmaketitle

    \acresetall
    
    \section{Introduction}
\label{sec:Introduction}


\IEEEPARstart{C}{limate} change has exposed an increasing population to flood risks, especially over the last decade\cite{Tellman2021SatelliteFloods}. 
Floods, even with low return periods, tend to cause damage that may take years to recover and these efforts tend to disfavor low-income, racial and ethnic minorities due to a difficulty in accessing federal resources as well as time taken for congressional appropriation \cite{Wilson2021}. With an average recurrence interval of at least once in 200 years, the Bellavue, TN flood event in 2021 tallied to over 100 million USD in costs and claimed at least 20 lives \cite{StormEventDatabaseEntry}. Similarly, with a return period of once in 400 years, large-scale flooding across Europe in 2021 added to the list of record-breaking floods in the region claiming approximately 240 lives and over 25 billion USD in damages\cite{ActurialPostEuropeanFloodDamageAssesmentArticle}.

Studying the dynamics of flood inundation can better equip stakeholders to mitigate the damage caused by such floods \cite{Romero2015RemoteSensingFloodInundation}. While \acp{WFM} -- which indicate the fraction of flood inundated pixels in the representative region -- are obtainable on a daily basis, their resolution may prove too coarse to be useful for tasks such as the study of inundation dynamics. On the other hand, fine-resolution \acp{FIM}, while useful for such tasks, are not available on a daily basis. Hence, the need arises to have more frequent, high-resolution \acp{FIM} which are binary images that indicate whether a specific location is inundated with water. This will aid in the study of inundation dynamics whose use-cases range from improved hydrological models, pre-flood mitigation strategies, underwriting flood insurance, property evaluation, and formulating evacuation plans to name a few.

Recent progress in satellite-based remote-sensing algorithms have provided stakeholders with flood related observations at global scales with existing products such as the NASA/NOAA Visible Infrared Imaging Radiometer Suite (VIIRS)\footnote{\href{https://www.nesdis.noaa.gov/current-satellite-missions/currently-flying/joint-polar-satellite-system/visible-infrared-imaging}{https://www.nesdis.noaa.gov/current-satellite-missions/currently-flying/joint-polar-satellite-system/visible-infrared-imaging}}, NASA Moderate Resolution Imaging Spectroradiometer(MODIS)\footnote{\url{https://modis.gsfc.nasa.gov}} and NASA/USGS Landsat\footnote{\href{https://landsat.gsfc.nasa.gov/}{https://landsat.gsfc.nasa.gov/}}. However, the coarse spatial resolution of MODIS (250m) and the long revisit times of  Landsat ($\sim$16 days) hinder the analysis of the spatiotemporal dynamics of flood hazard at a range of spatial scales (from small creeks to big rivers) and terrains (from natural floodplains to urban settlements). \acp{WFM}, are available via existing satellite products (NOAA VIIRS and NASA MODIS) at a daily temporal frequency and at the cost of reduced image resolution. Naturally, using these \acp{WFM} to produce high quality \acp{FIM} at higher spatio-temporal resolution can prove advantageous to studying flood dynamics at a finer scales.

In this work, we investigate the task of downscaling low-resolution \acp{WFM} (300m) to high-resolution \acp{FIM} (30m) with computational approaches inspired by the successes of deep learning models in the field of super-resolution imaging \cite{Wang2021DeepSurvey}. We explore the utility of three state-of-the-art deep learning models --namely \ac{RCAN}, \ac{RDN} and \ac{ESRT}. The circumstances of our problem setting introduce a training data scarcity problem due to the following reasons: (i) we can record high resolution ground truth \acp{FIM} from Landsat only once every 16 days, (ii) the probability of observing a flood event during these satellite visits further reduces our opportunities to collect data and (iii) flood events are relatively rare. These constraints makes it extremely hard to compile high quality datasets for a region of interest, especially for the data-driven models we seek to train. To alleviate this problem, we opted to use \ac{SYN} data generated by physics-based simulations wherein the resolution can be controlled, admittedly at a significantly higher computational cost. With this solution, we postulate that \ac{SYN} data can function as a viable proxy for \ac{RW} data.

We conducted experiments on four regions -- namely, Iowa, Western Europe, Red-River and Ghana -- and show that:

\begin{itemize}
    \item \ac{SYN} data are a viable proxy to existing scarce \ac{RW} flood inundation data.
    \item Among the models we've trained using the \ac{SYN} Iowa data, there are performance benefits, when evaluated on \ac{RW} Iowa data.
    \item Our trained ML models are transferable with zero-shot performance benefits when applied to hydroclimatologically similar regions (according the K\"{o}ppen classification scheme \cite{Chen2013GeoMorphologicalSimilarity}) such as Western Europe and fails to show significant benefit when evaluated in hydroclimatologically dissimilar regions such as Ghana and Red-River. 
\end{itemize}

The rest of the paper is organized as follows. Section \ref{sec:RelatedWork} details some of the existing works and their limitations as we pave the path towards our proposed models in Section \ref{sec:Methodology}. Section \ref{sec:DataDescription} describes the datasets and pre-processing steps that we employed for this study. We describe our experimental setting in Section \ref{sec:expriments}, followed by a discussion of the results in Section \ref{sec:Results}. The GitHub repository containing the code and data has also been published\footnote{\url{https://github.com/aaravamudan2014/SIDDIS}}. 


    \section{Related Work}
\label{sec:RelatedWork}

Within the realm of geophysical sciences, super-resolution/downscaling is a challenge that scientists continue to tackle. There have been several works involved in downscaling applications such as river mapping \cite{Yin2022}, coastal risk assessment \cite{Rucker2021}, estimating soil moisture from remotely sensed images \cite{Peng2017SoilMoisture} and downscaling of satellite based precipitation estimates \cite{Medrano2023PrecipitationDownscaling} to name a few. We direct the reader to \cite{Karwowska2022SuperResolutionSurvey} for a comprehensive review of satellite based downscaling applications and methods. Pertaining to our objective of downscaling \acp{WFM}, we can draw comparisons with several existing works. 
In what follows, we provide a brief review of functionally adjacent works to contrast the novelty of our proposed model and its role in addressing gaps in literature. 

When it comes to downscaling \ac{WFM}, several works use statistical downscaling techniques. These works downscale images by using statistical techniques that utilize relationships between neighboring water fraction pixels. For instance, \cite{Li2015SRFIM} treat the super-resolution task as a sub-pixel mapping problem, wherein the input fraction of inundated pixels must be exactly mapped to the output patch of inundated pixels. 
\cite{Wang2019} improved upon these approaches by including a spectral term to fully utilize spectral information from multi spectral remote sensing image band. \cite{Wang2021} on the other hand also include a spectral correlation term to reduce the influence of linear and non-linear imaging conditions. All of these approaches are applied to water fraction obtained via spectral unmixing \cite{wang2013SpectralUnmixing} and are designed to work with multi spectral information from MODIS. However, we develop our model with the intention to be used with water fractions directly derived from the output of satellites. One such example is NOAA/VIIRS whose water fraction extraction method is described in \cite{Li2013VIIRSWFM}. \cite{Li2022VIIRSDownscaling} presented a work wherein \ac{WFM} at 375-m flood products from VIIRS were downscaled 30-m flood event and depth products by expressing the inundation mechanism as a function of the \ac{DEM}-based water area and the VIIRS water area.

On the other hand, the non-linear nature of the mapping task lends itself to the use of neural networks. Several models have been adapted from traditional single image digital super-resolution in computer vision literature \cite{sdraka2022DL4downscalingRemoteSensing}. Existing deep learning models in single image super-resolution are primarily dominated by \ac{CNN} based models. Specifically, there has been an upward trend in residual learning models. \acp{RDN} \cite{Zhang2018ResidualDenseSuperResolution} introduced residual dense blocks that employed a contiguous memory mechanism that aimed to overcome the inability of very deep \acp{CNN} to make full use of hierarchical features. 
\acp{RCAN} \cite{Zhang2018RCANSuperResolution} introduced an attention mechanism to exploit the inter-channel dependencies in the intermediate feature transformations. There have also been some works that aim to produce more lightweight \ac{CNN}-based architectures \cite{Zheng2019IMDN,Xiaotong2020LatticeNET}. Since the introduction of the vision transformer \cite{Vaswani2017Attention} that utilized the self-attention mechanism -- originally used for modeling text sequences -- by feeding a sequence 2D sub-image extracted from the original image. Using this approach \cite{LuESRT2022} developed a light-weight and efficient transformer based approach for single image super-resolution.

For the task of super-resolution of \acp{WFM}, we discuss some works whose methodology is similar to ours even though they differ in their problem setting. \cite{Yin2022} presented a cascaded spectral spatial model for super-resolution of MODIS imagery with a scaling factor 10. Their architecture consists of two stages; first multi-spectral MODIS imagery is converted into a low-resolution \ac{WFM} via spectral unmixing by passing it through a deep stacked residual \ac{CNN}. The second stage involved the super-resolution mapping of these \acp{WFM} using a nested multi-level \ac{CNN} model. Similar to our work, the input fraction images are obtained with zero errors which may not be reflective of reality since there tends to be sensor noise, the spatial distribution of whom cannot be easily estimated. We also note that none of these works directly tackle flood inundation since they've been trained with river map data during non-flood circumstance and \textit{ergo} do not face a data scarcity problem as we do. 
\cite{Jia2019} used a deep \ac{CNN} for land mapping that consists of several classes such as building, low vegetation, background and trees. 
\cite{Kumar2021} similarly employ a \ac{CNN} based model for downscaling of summer monsoon rainfall data over the Indian subcontinent. Their proposed Super-Resolution Convolutional Neural Network (SRCNN) has a downscaling factor of 4. 
\cite{Shang2022} on the other hand, proposed a super-resolution mapping technique using Generative Adversarial Networks (GANs). They first generate high resolution fractional images, somewhat analogous to our \ac{WFM}, and are then mapped to categorical land cover maps involving forest, urban, agriculture and water classes. 
\cite{Qin2020} interestingly approach lake area super-resolution for Landsat and MODIS data as an unsupervised problem using a \ac{CNN} and are able to extend to other scaling factors. \cite{AristizabalInundationMapping2020} performed flood inundation mapping using \ac{SAR} data obtained from Sentinel-1. They showed that \ac{DEM}-based features helped to improve \ac{SAR}-based predictions for quadratic discriminant analysis, support vector machines and k-nearest neighbor classifiers. While almost all of the aforementioned works can be adapted to our task. We stand out in the following ways (i) We focus on downscaling of \acp{WFM} directly, \textit{i.e.,} we do not focus on the algorithm to compute the \ac{WFM} from multi-channel satellite data and (ii) We focus on producing high resolution maps only for instances of flood inundation. The latter point produces a data scarcity issue which we seek to remedy with synthetic data.

    \section{Data Description}
\label{sec:DataDescription}

\subsection{Synthetic \& Real-world Data}
We used synthetic flood inundation data for the entire state of Iowa, provided by the Iowa Flood Center \cite{RealTimeFloodForecastingandInformationSystemfortheStateofIowa2017}. Synthetic flood maps were derived from steady-state simulations for design flows that correspond to a range of probability of exceedance (USGS, Bulletin 17) using one-dimensional open-channel flow model from the Hydrologic Engineering Center’s River Analysis System (HEC-RAS) 5. The river channel geometric properties, such as slopes
and cross sections, were derived from airborne LiDAR based one-meter \ac{DEM} data. For more details of the flood inundation mapping procedures see \cite{Gilles2012InundationMappingInitiativesIowaFloodCenter}.
These simulated \acp{FIM} counter the issue of securing an already-limited availability of real-world \acp{FIM}, and proves advantageous to deep learning models that leverage increased data input for the super resolution task. 

\begin{figure*}[ht!]
\includegraphics[width=\textwidth, angle=0]{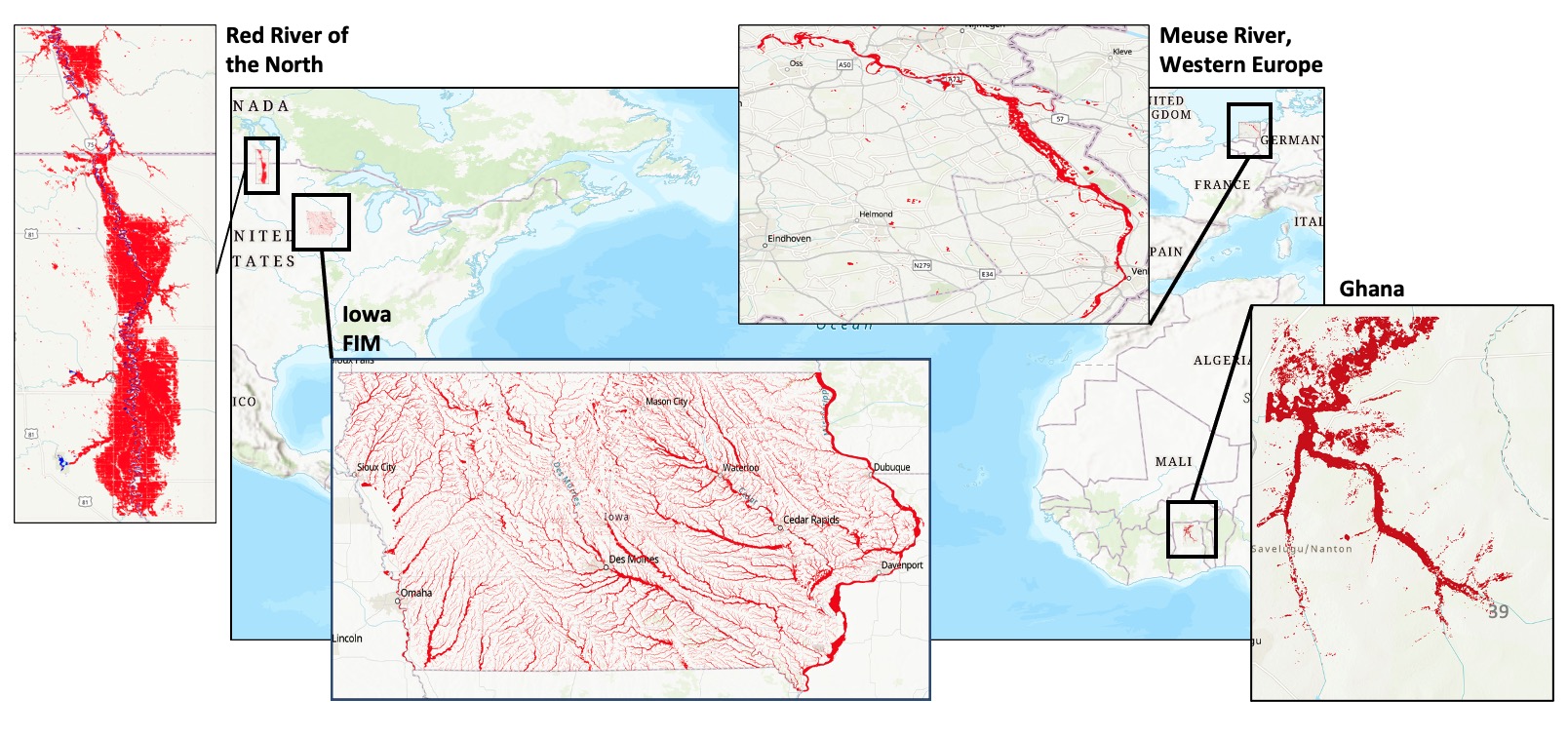}
\caption{ Chosen regions for this study. Iowa is the region where the model was trained over, whereas all the other regions comprise of climatically similar (Europe) and dissimilar (Ghana, Red River) regions. \label{fig:DatasetLocations}}
\end{figure*}

The test dataset contained a withheld sample of the synthetic \acp{FIM} but, more importantly, we prepared \ac{RW}, satellite-based data derived from Landsat-8 output. These \ac{RW} samples allow us to evaluate the generalizability of our trained downscaling model. Our targeted test regions utilizing \ac{RW} data were located both in and out of Iowa (the region used for training). This motivates our research question about how effective deep learning models can transfer across other hydroclimatological regions. Landsat-8-based \acp{FIM} were generated for two sites within Iowa: at Cedar River and Des Moines River. Outside of the training region, \acp{FIM} were prepared for the flood events occurring in Red River in the North of the USA, the Nasia river in Ghana and Western Europe. \tabref{tab:dataset_locations} contains the chosen abbreviations for each of these study regions along with the period when the flooding occurred. Figure \ref{fig:DatasetLocations} shows the chosen locations for this paper. 

\textbf{Generating \acp{WFM}:} For both \ac{SYN} and \ac{RW} data, we produce the \acp{WFM} by finding the ratio of inundated pixels in every 10x10 patch of pixels in the high resolution \ac{FIM}. For \ac{RW} data, we can obtain coarse resolution \ac{RW} \acp{WFM} from existing satellite products such as NOAA/VIIRS and the equivalent high resolution \ac{FIM} from other satellite observations such as Landsat. However, in reality the differences between the two in terms of, primarily, timing of observations and secondly the algorithms applied for water detection, introduce discrepancies between coarse and high-res scenes that would contaminate the evaluation of high-res \acp{FIM} produced by the super-resolution algorithms. To avoid this source of error and focus the evaluation only on the effectiveness of the algorithms examined, we elected to create the coarse resolution \ac{RW} \acp{WFM} by aggregating the high-res \acp{FIM} from Landsat. This means that, by construction, the number of inundated pixels in the high resolution \ac{FIM} is exactly obeyed. However this may not be reflective of reality since real world \ac{WFM} may be inundated with sensor noise, along with the errors induced by the algorithm that generate these \acp{WFM}.

\subsection{Pre-processing coarse- and high-resolution imagery}

\ac{RW} \acp{FIM} were generated using multi-spectral images of Landsat-8 for the aforementioned flood events. Cloud free scenes were selected and the \ac{SWI}, defined as $\mathrm{SWI} \triangleq \frac{Blue - \mathrm{NIR}} { Red+Green+Blue}$,  was used for water detection, where NIR refers to the Near Infrared band. An \ac{SWI} threshold value was selected for each flood event after quantitative comparison of the histograms between non-flooded and flooded scenes and was applied for the binary classification of pixels to flooded and non-flooded. The resulting \acp{FIM} were also visually compared against visible Landsat-8 images for consistency. The coarse resolution \acp{WFM} that provide pixel values recorded as water fractions in the range $[0, 1]$ were prepared when $30m$ high-resolution \acp{FIM} having frame sizes $100 \times 100$ were down-sampled by a scale factor of 10. Resultantly, coarse resolution \acp{FIM} were $300m$/pixel, $10 \times 10$ images. 
The high resolution \acp{FIM} reported binary pixel values of 0 or 1 indicating no-flood and flood cells, respectively. 


\begin{table*}
\caption{Chosen regions of this study\label{tab:dataset_locations}. We group the datasets into Iowa and External datasets. }
\centering
\begin{tabular}{lllll}
\hline
\textbf{Training FIMs  {[}Location{]}} & \textbf{Validation/Test FIMs  {[}Location{]}}            & \textbf{Time period of flood.} & \textbf{Abbrev.} & \textbf{Samples} \\ \hline
\multirow{3}{*}{{[}Iowa{]}} & Synthetic      {[}Iowa{]}                                 & N/A  & SYN-IA  & 2,610\\ 
                                       & Landsat-8     {[}Cedar River, Iowa{]}& Sep 2016 & RW-IA(CR)   &  583\\ 
                                       & Landsat-8     {[}Des Moines River, Iowa{]} & March 2019 & RW-IA(DM)   &  531\\\cline{1-5} 
\multirow{3}{*}{{[}External{]}}
                                       & Landsat-8      {[}Europe{]}                            &July 2021   & RW-EU   & 311\\ 
                                       & Landsat-8      {[}Red River, North America{]}                               & April 2020 &RW-RR  & 582 \\ 
                                       & Landsat-8      {[}Nasia River, Ghana{]}                               & Sep 2007 & RW-GH  & 31 \\ \hline

\end{tabular}

\end{table*}

\section{Methodology}
\label{sec:Methodology}

We denote a low-resolution \ac{WFM} by $\mathbf{X} \in [0, 1]^{L \times L}$. Our objective will be to learn a map $\mathbf{M}: [0, 1]^{L \times L} \to \{0, 1\}^{H \times H}$ that will downscale $\mathbf{X}$ to a high-resolution \ac{FIM} $\hat{\mathbf{Y}} \in \{0, 1\}^{H \times H}$, where $H = fL$ and $f > 1$ is a scale factor, so that $\hat{\mathbf{Y}}$ recovers the (true) \ac{FIM} $\mathbf{Y}$ that is associated to $\mathbf{X}$. In our work, we use $f=10$ to replicate the spatial resolution produced by existing high resolution satellite products. In order to learn $\mathbf{M}(\cdot)$, we assume that each model we consider aims to learn the conditional joint probability $p(\mathbf{Y} | \mathbf{X})$. Furthermore, for simplicity, we will assume that the entries of $\mathbf{Y}$ (\ac{FIM} pixel intensities) are mutually independent, when conditioned on the entries of $\mathbf{X}$ (\ac{WFM} pixel intensities), \textit{i.e.}, $p(\mathbf{Y} | \mathbf{X}) = \prod_{i,j} p(Y_{i,j} | \mathbf{X})$. Finally, assuming that the model accurately estimates the probabilities $p(Y_{i,j} = 1 | \mathbf{X})$ via $S_{i,j}(\mathbf{X}) \in [0,1]$, the \ac{FIM} pixel intensities are reconstructed as $\hat{Y}_{i,j} = 1$, if $S_{i,j}(\mathbf{X}) \geq 0.5$ and $0$, if otherwise. In what follows, we present the neural-based architectures we have chosen to implement the map $\mathbf{M}(\cdot)$ for this downscaling task.


\subsection{Model Architectures}
\label{subsec:ModelArchitecture}


There have been a variety of works that use deep learning techniques to learn highly complex mappings between the low- and high-resolution images. Note that, in our data, most of regions are not inundated with floods. This means that the output patch that corresponds to a water fraction of zero will be all zeros (no inundation). The same applies to fully inundated pixels. Therefore, once mapped to a higher resolution, we simply need to learn the residual map between the \ac{WFM} and \ac{FIM}. A residual connection from the input to the output therefore plays an important role in such architectures. This connection, in essence, crudely downscales a \ac{WFM} to a \ac{FIM} and the network learns a modification to this mapping that performs better at downscaling. This naturally lead us to the path of residual learning. Residual learning aims to learn the residual, or difference, between the output and input images \cite{He2016DeepRecognition}.  Residual learning is mainly motivated by the following reasons (i) an unexpected training performance degradation, when networks grow deeper and, hence, should overfit and perform better and (ii) training becomes less prone to exploding or vanishing gradients. 


In order to be able to train deep architectures we will rely on residual learning, which offer the aforementioned benefits. More specifically, we use \acp{RDN} \cite{Zhang2018ResidualDenseSuperResolution} and \acp{RCAN} \cite{Zhang2018RCANSuperResolution} (see \cite{Wang2021DeepSurvey} for a comparative analysis).
Apart from this, we also consider an efficient transformer based architecture called \ac{ESRT} \cite{LuESRT2022} that has been shown to be a strong competitor of late. A common thread among all of these architectures is a shallow feature extraction layer and a final dense fusion layer. The final stage incorporates global residual learning and feature fusion to produce feature vectors, which are then passed through a transposed convolution layer  -- see  \cite{Shi2016SubPixelConvolution} -- in order to bring it to a size of $100 \times 100$ before being passed through a few more convolution layers to produce the high resolution \ac{FIM}. In the following subsections, we briefly describe the individual architectures.

\subsubsection{RDN}

\acp{RDN} \cite{Zhang2018ResidualDenseSuperResolution} combine the use of residual learning and of densely connected convolutional blocks: each such block consists of a number of convolutional layers, whose inputs consist of the outputs of all previous layers (via skip connections) within the same block; these inputs are regarded as hierarchical features. Finally, a skip connection routes the input of such blocks to their outputs in order to implement a type of local residual learning. \acp{RDN} employ global residual learning and consist of a long cascade of such blocks and a final upsampling stage that yields the downscaled image. In our \ac{RDN} architecture, we use 12 features in the convolutional layers, a kernel size of 3, 20 residual blocks and 20 layers per residual block.



\subsubsection{RCAN}

The \ac{RCAN} was introduced in \cite{Zhang2018RCANSuperResolution}. \acp{RCAN} make use of channel attention, which aims to exploit the inter-dependencies between feature channels. This is done by first aggregating channel-wise features and applying a gating mechanism that learns non-linear relationships between the feature channels. Following this, the features are passed through Residual-in-Residual (RIR) blocks, wherein residual learning is enforced. RIR blocks contain \acp{RCAB}, which allows the network to focus on the important aspects of the low resolution features. In our model, we used 10 residual groups, each of which contains several residual blocks with short skip connection, 10 residual channel attention blocks, 64 features, kernel size of 3 and a reduction factor of 16.



\subsubsection{ESRT}

Transformers have shown to be quite effective in sequential tasks  for natural language processing in comparison to convolutional neural networks \cite{Vaswani2017Attention}. The self-attention mechanism in transformers was applied to computer vision tasks in \cite{dosovitskiy2021an} by treating each image as a sequence of sub-images. While transformers typically feature an encoder and a decoder block when used for natural language processing tasks, for computer vision tasks they only feature an encoder to embed images into some feature space and then use another deep learning architecture for downstream tasks such as classification and super-resolution. \cite{LuESRT2022} proposed the \acf{ESRT}, a low-complexity transformer architecture tailored to super-resolution tasks. In our usage of \ac{ESRT}, we used 3 encoder layers with 32 features and kernel size of 3, each with a channel attention layer. The attention layer was a multi-scale local attention block consisting of 288 features.

\subsection{Training}
\label{subsec:training}

All of our data-driven, downscaling models were trained to minimize a penalized version of 
the average (over pixels) cross-entropy for each data pair $(\mathbf{X}, \mathbf{Y})$
\begin{align}
    L_{\mathrm{PACE}}(\mathbf{X}, \mathbf{Y}) & \triangleq L_{\mathrm{ACE}}(\mathbf{Y}, \mathbf{S}(\mathbf{X})) + \eta 
 P(\mathbf{X}, \mathbf{S}(\mathbf{X}))   
    \shortintertext{where $\eta \geq 0$ is a penalty parameter and}
    L_{\mathrm{ACE}}(\mathbf{Y}, \mathbf{S}(\mathbf{X}))  & \triangleq \frac{1}{H^2} \sum_{i,j} \left[ Y_{i,j} \ln S_{i,j}(\mathbf{X}) + \right. \nonumber
    \\
    & \left. + (1 - Y_{i,j}) \ln (1 - S_{i,j}(\mathbf{X})) \right]
    \\
    P(\mathbf{X}, \mathbf{S}(\mathbf{X})) & \triangleq \sum_{i,j} \left( X_{i,j} - \frac{1}{f^2} \sum_{k,l} S_{fi+k, fj+l}(\mathbf{X}) \right)^2
\end{align}
where $i,j \in \{0, 1, \ldots, L-1\}$ and $k,l \in \{0, 1, \ldots, f-1\}$. The penalty term $P(\cdot, \cdot)$ quantifies the deviation between the fraction of inundated pixels in a $f \times f$ \ac{FIM} patch vis-\'{a}-vis its corresponding \ac{WFM} pixel. Using higher values of the penalty parameter $\eta$ for training our downscaling models, significantly penalizes even modest deviations and, in essence, enforces the matching of water fractions between \acp{FIM} and \acp{WFM}. In our study, this was important as such matching occurred in our \ac{SYN} data. Finally, let us point out that we treated $\eta$ as a model hyper-parameter.





\begin{figure*}[htpb!]
  \centering
  \captionsetup{justification=centering} \includegraphics[width=\textwidth,height=15cm]{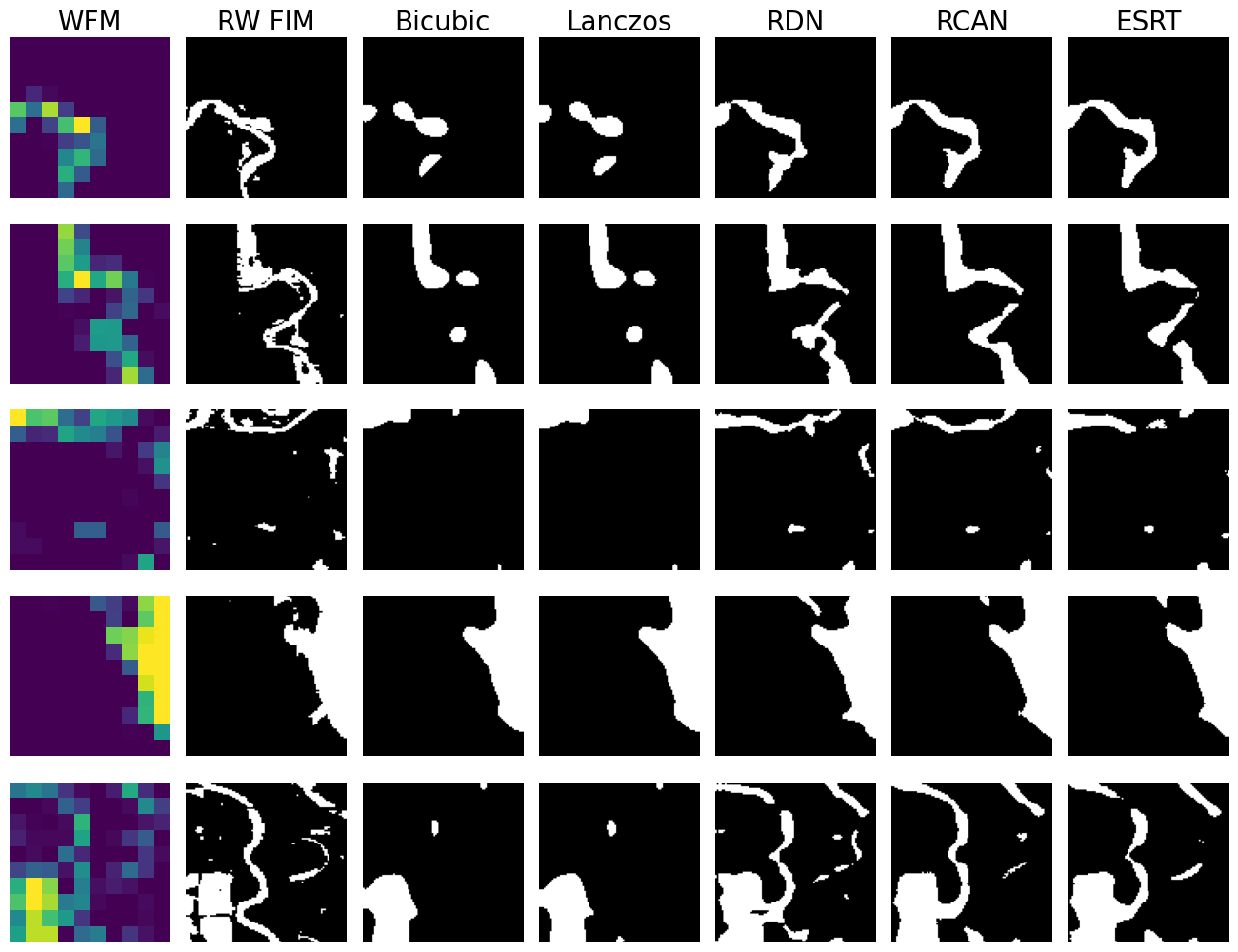}
  \caption{\label{fig:RWIowaDesMoines} Sample outputs for the \ac{RW} Iowa Des Moines region; \textit{i.e.,} the region over which the model was trained. }
\end{figure*}

\begin{figure*}[htpb!]
  \centering
  \captionsetup{justification=centering}
  \includegraphics[width=\textwidth,height=15cm]{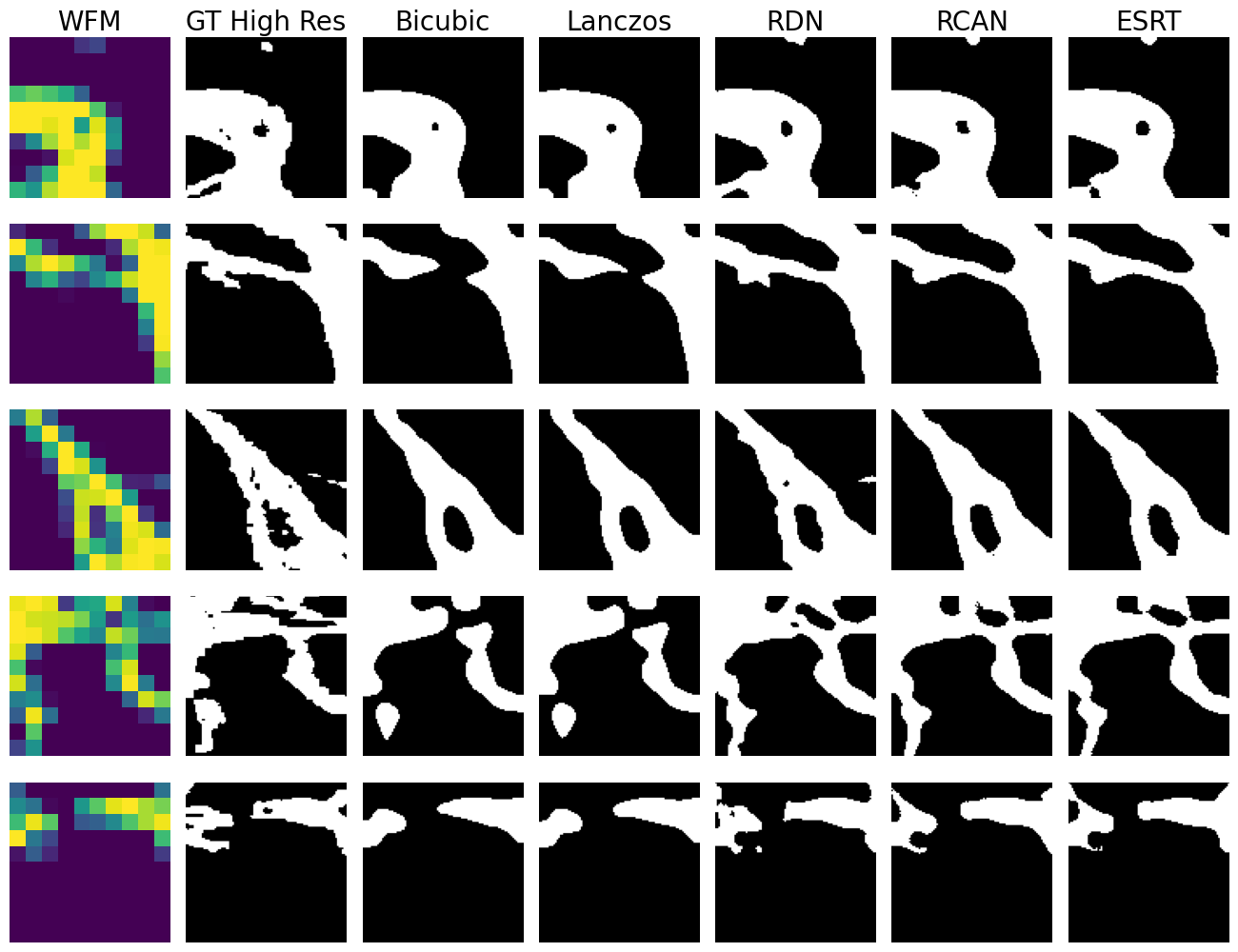}
  \caption{\label{fig:EUSamples} Sample outputs for the \ac{RW} EU region; \textit{i.e.,} external to the regions over which the model was trained.}
\end{figure*}

The weights of the neural network were optimized using Adam \cite{Kingma2015} with a decaying learning rate. The hyper-parameters along with their ranges in parentheses are as follows. The learning rate ($10^{-5}$ to $10^{-4}$), $\eta$ (0 to 2000), layer dropout probability ($10^{-3}$ to $20^{-2}$), random seed (100 to 900 in multiples of 100). For the \ac{RDN}, number of features (8 to 64 in mutliples of 8), number of residual blocks (2 to 16 in multiples of 2), number of layers per block (2 to 32 in multiples of 2) and the kernel size was fixed to 3. For the \ac{RCAN}, number of features (4 to 64 in multiples of 4), number of residual groups (10 to 40 in multiples of 5), number of residual channel attention blocks (20 to 50 in multiples of 5), reduction factor (16 to 64 in multiples of 4). For the \ac{ESRT}, we used the hyper-parameters as available in the code for \cite{LuESRT2022}. We used Optuna\footnote{\href{https://optuna.readthedocs.io/en/stable/\#}{https://optuna.readthedocs.io/en/stable/\#}} which in turn used the Tree-structured Parzen Estimator \cite{Bergstra2011HPOptimizationAlgorithms} to produce candidates for hyper-parameter search.

\begin{figure*}[htpb!]
  \centering
  \captionsetup{justification=centering}
  \includegraphics[width=\textwidth,height=15cm]{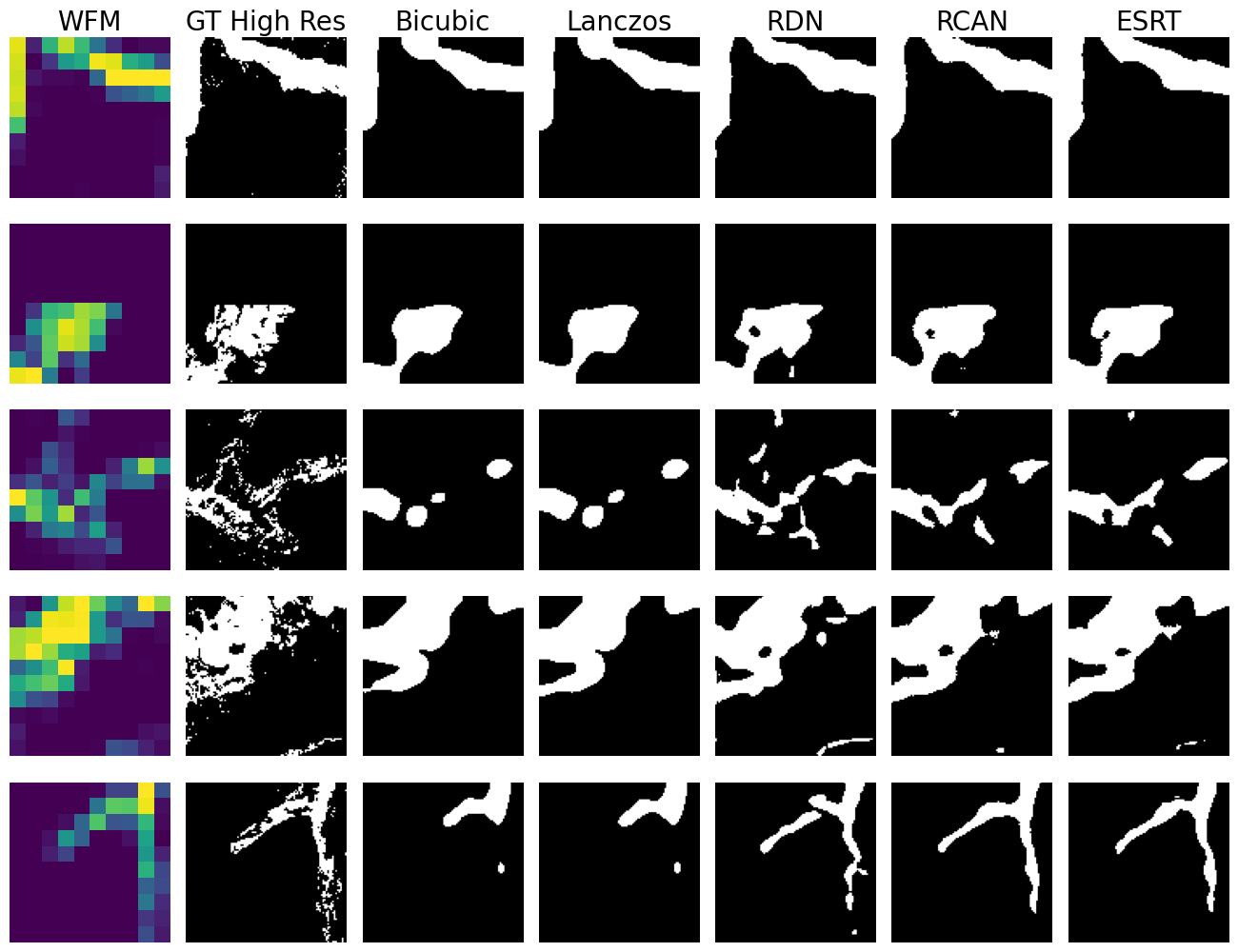}
  \caption{\label{fig:GhanaSamples} Sample outputs for the \ac{RW} Ghana region; \textit{i.e.,} external to the regions over which the model was trained.}
\end{figure*}




    \section{Experimental Evaluation}
\label{sec:expriments}

\begin{table*}
\centering
\caption{\label{tab:IowaResults}Accuracies, along with their Prediction Intervals (PI) and \Acf{MCC} of our models for Iowan \Acf{RW} and \Acf{SYN} data. \Acf{RW} were obtained from Landsat-8 data, while \acf{SYN} data were provided by the Iowa Flood Center. Note that we only present results for pixels where the water fraction in the \ac{WFM} is in the interval (0.25, 0.85). Na\"ive model here represents outputs where all the pixels are predicted as non-inundated, \textit{i.e.,} the majority class. Percentages depicted in bold signify the best performing model in each case.}
\vspace*{0.25cm}
\begin{tabular}{@{}cccccccccc@{}}
\toprule
        & \multicolumn{3}{c}{SYN-Iowa}                       & \multicolumn{3}{c}{RW-Iowa (CR)}                  & \multicolumn{3}{c}{RW-Iowa (DM)}                  \\ \midrule
Model   & accuracy (\%)  & accuracy PI (\%) & MCC                        & accuracy(\%)  & accuracy PI (\%) & MCC                    & accuracy(\%)  & accuracy PI (\%) & MCC                 \\\midrule
Na\"ive & 50.74   & [50.66, 50.82]       & 0.0             & 49.83    & [49.55, 50.10]      & 0.0            & 52.29   & [52.02, 52.56]       & 0.0                  \\
bicubic & 73.27   & [73.20, 73.35]       & 0.475             & 69.96    & [69.71, 70.21]      & 0.404            & 70.63    & [70.38, 70.87]      & 0.418                  \\
Lanczos & 73.72   & [73.65, 73.79]       & 0.483               & 70.25    & [70.00, 70.49]      & 0.409                  & 70.89   & [70.65, 71.14]       & 0.423               \\
RDN     & 78.64   & [78.57,78.70]       & 0.574         \      & 70.65    & [70.40, 70.90]      & 0.414          & 70.52            & [70.28, 70.77]       & 0.410               \\
RCAN    & 79.91   & [79.85, 79.98]       & 0.599                  & 72.69    & [72.45, 72.93]      & 0.454                  & 72.93    & [72.69, 73.17]      & 0.457             \\
ESRT    & \textbf{80.33} & \textbf{[80.27, 80.40]} & \textbf{0.607}  & \textbf{73.31} & \textbf{[73.07, 73.55]} & \textbf{0.466} & \textbf{73.34} & \textbf{[73.10, 73.58]} & \textbf{0.465}  \\ \bottomrule
\end{tabular}
\end{table*}

\begin{table*}[!ht]
\vspace*{0.52cm}
\centering
\caption{\label{tab:ExternalResults} Accuracies along with their Prediction Intervals (PI) and \Acf{MCC} of our models for all external \Acf{RW} data. \Acf{RW} were obtained from Landsat-8 data. Note that we only present results for pixels where the water fraction in the \ac{WFM} is in the interval (0.25, 0.85). Na\"ive model here represents outputs where all the pixels are predicted as non-inundated, \textit{i.e.,} the majority class. Percentages depicted in bold signify the best performing model in each case.}
\vspace*{0.25cm}
\begin{tabular}{@{}cccccccccc@{}}
\toprule
        & \multicolumn{3}{c}{RW-EU}                         & \multicolumn{3}{c}{RW-GH}                        & \multicolumn{3}{c}{RW-RR}                          \\ \midrule
Model   & accuracy (\%) & accuracy PI (\%) & MCC                     & accuracy(\%)  & accuracy PI (\%) & MCC                        & accuracy(\%) & accuracy PI (\%)  & MCC                          \\ \midrule
Na\"ive & 47.03    & [46.67, 47.39]      & 0.0             & 47.77    & [47.08, 48.46]      & 0.0            & 40.86     &   [40.71, 41.01]   & 0.0                  \\
bicubic & 77.77    & [77.47, 78.07]      & 0.558                & 73.11    & [72.50, 73.73]      & 0.463                  & 71.39    &  [71.25, 71.52]      & 0.395             \\
Lanczos & 78.38    & [78.09, 78.68]      & 0.570                   & 73.50     & [72.89, 74.11]     & 0.471                  & 71.55   & [71.42, 71.69]       & 0.399             \\
RDN     & 80.97    & [80.69, 81.25]      & 0.618                   & 71.70     & [71.07,72.32]     & 0.432                   & 69.60    & [69.46, 69.74]       & 0.361           \\
RCAN    & 82.13    & [81.86, 82.41]      & 0.641                  & \textbf{73.97}     & [73.36, 74.57]     & \textbf{0.478}                & 71.89   & [71.76, 72.03]        & 0.4083                   \\
ESRT    & \textbf{83.27} & \textbf{[83.00, 83.54]}& \textbf{0.664}  &  73.72 & [73.11, 74.57]& 0.474  & \textbf{72.32} &\textbf{[72.18, 72.45]} & \textbf{0.419}  \\ \bottomrule
\end{tabular}
\end{table*}

\subsection{Baseline Algorithms} 

We compared our neural-based downscaling models to two baseline methods, namely bicubic and Lanczos interpolation. Both of these methods are extensively used in common image processing tasks, including image downscaling. The former uses a polynomial kernel, while the later uses a product of cardinal sines to interpolate between \ac{WFM} samples. The resulting image intensities are subsequently thresholded to yield a binary-valued \ac{FIM}. 


\subsection{Metrics} 
All the downscaling models we consider employ an adjustable threshold $\theta \in [0,1]$, based on which an array of pixel-on probabilities $S_{i,j}$ is thresholded to obtain a binary \ac{FIM}. We will refer to the fraction of \ac{FIM} pixels whose state (inundated \textit{vs.} non-inundated) is correctly predicted, when using $\theta = 0.5$, as \textit{accuracy}. Apart from this metric and due to the asymmetric importance of predicting inundation \textit{vs.} predicting the lack of it at a locality, we also recorded the \ac{ROC} curve for each model, which depicts the model's true positive rate (hit rate) as a function of the false positive rate (false alarm rate). These curves were obtained by varying the threshold $\theta$ between $0$ and $1$ to obtain predictions from the trained models and allow a stakeholder to establish an acceptable false alarm rate. Finally, due to the pronounced imbalance between the numbers of inundated and dry localities, we also report the models' \acf{MCC}, which ranges in $[0,1]$ and gauges how much more accurate a model is over always predicting that every locality is dry. An \ac{MCC} of $0$ indicates no improvement over a na\"ive model that always predicts no inundation for all pixels. Finally, we need to note that we only aggregate the results for pixels wherein the water fraction is between $0.25$ and $0.85$. This was done to exclude non-riverine regions and the areas in the middle of the river that contribute heavily to the aforementioned metrics due to their large proportion in \acp{FIM}. We also report the Prediction Intervals (PI) of the accuracies for each model. This was calculated using the Clopper-Pearson \cite{ClopperPearsonPI1934} prediction intervals for binomial proportion with a confidence level of 0.99.

\section{Results and Discussions}
\label{sec:Results}

In this section, we aim to answer three main questions. First, we want to validate our hypothesis that \ac{SYN} data is a viable proxy for \ac{RW} data when training ML models for downscaling. Secondly, we seek to assess how much more skillful ML-based downscaling is compared to classical, non-data-driven techniques, such as our baseline methods, \textit{i.e.}, thresholded bicubic and Lanczos interpolation. Finally, we would like to appraise the extent to which data-driven models like ours are transferable (in terms of usefulness) to other regions without major performance degradations.  
To assess the quality of the models, we conduct a multiple comparison test --namely the Holm-Bonferroni procedure \cite{HolmBonferroni1979} -- that is designed to control the \ac{FWER}. We notice that, with a \ac{FWER} of $10^{-3}$, all the differences in model performance are significant. The only exception to this trend was observed in \ac{RW}-GH for whom the pairwise differences between \ac{RCAN} and \ac{ESRT}, Lanczos and Bicubic were not significant with the aforementioned \ac{FWER}. 


\subsection{Potential of using SYN Data for RW downscaling}

\begin{figure*}[t!]
    \centering
    \begin{subfigure}[t]{0.5\textwidth}
        \centering
        \includegraphics[width=\textwidth/2]{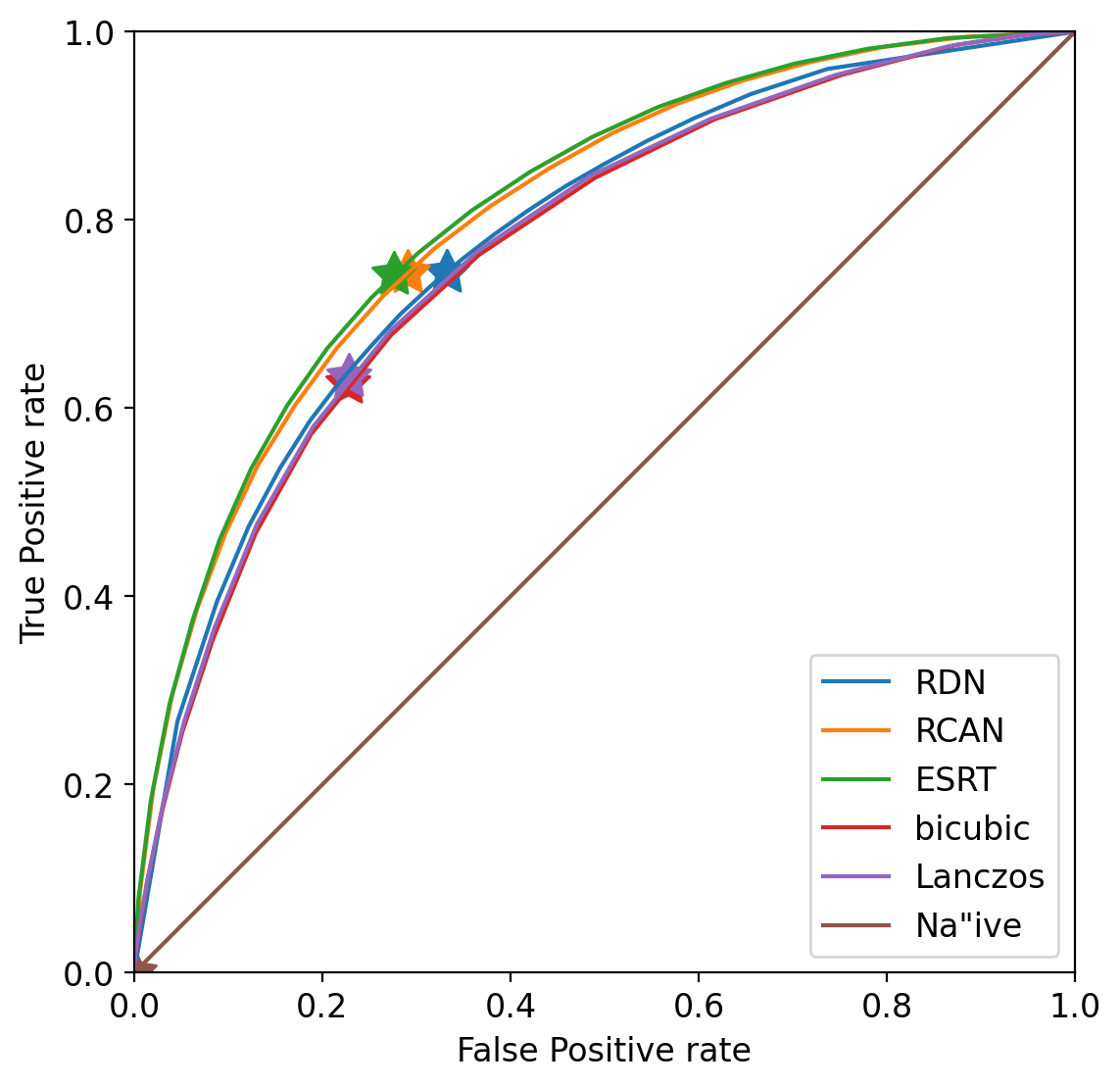}
        \caption{}
    \end{subfigure}%
    ~ 
    \begin{subfigure}[t]{0.5\textwidth}
        \centering
        \includegraphics[width=\textwidth/2]{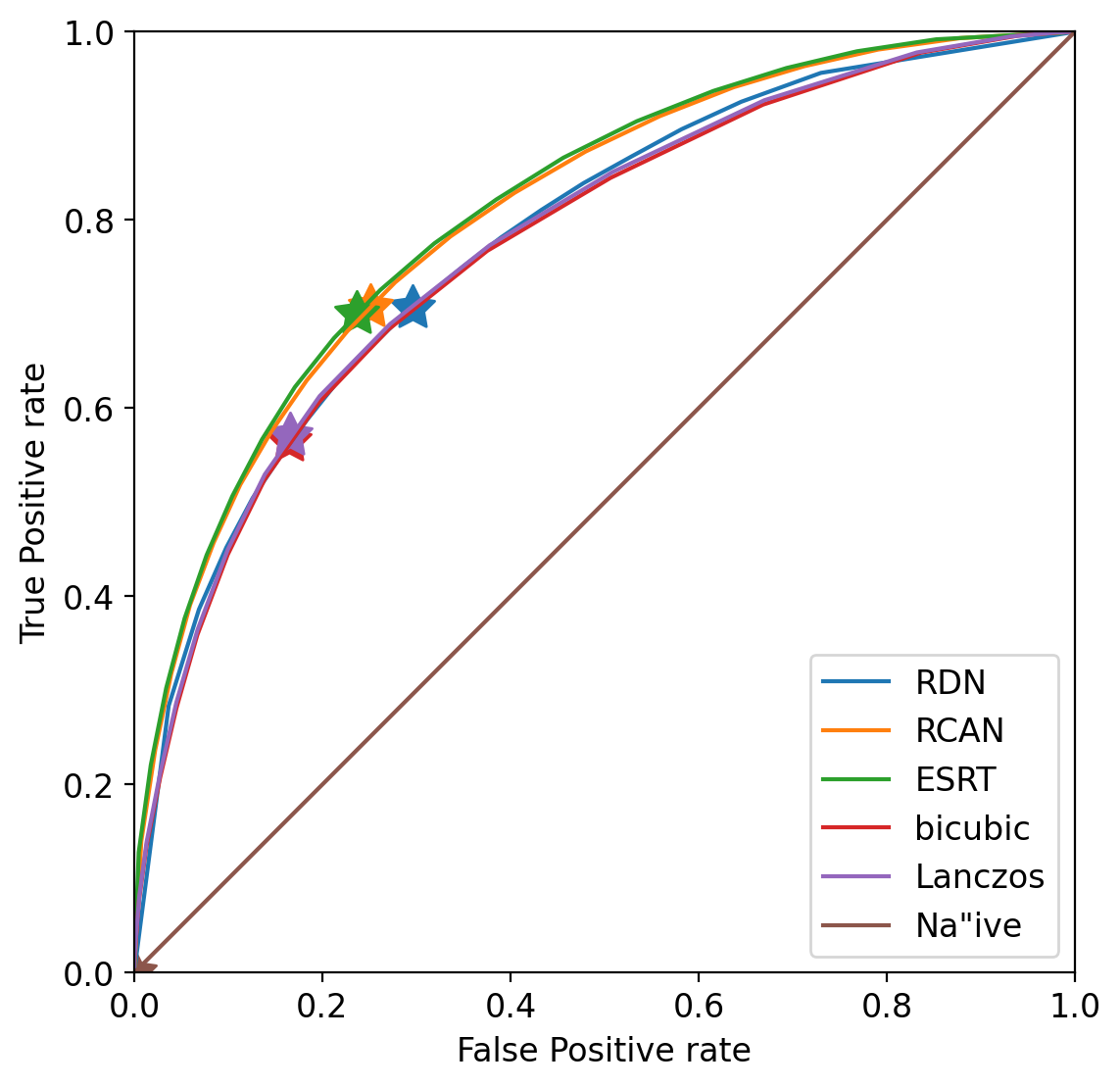}
        \caption{}
    \end{subfigure}
    \vspace*{0.5cm}
    \caption{    \label{fig:RW_ROC_Curves} \Acf{ROC} curves for (a) RW-IA (CR) and (b) RW-IA (DM) dataset. Na\"ive model here represents a model whose output is solely a ``no Flood'' for all pixels. Star here represents the pixel-wise classifier with a threshold of 0.5.}
\end{figure*}

In order to evaluate the utility of synthetic data for training, we compare performances of our candidate models on both \ac{SYN} and \ac{RW} Iowa data whose results are presented in Table \ref{tab:IowaResults}. We notice that 
\textbf{(i)} For the Iowa datasets, there is a drop in performance of all the models when going from \ac{SYN} to \ac{RW} datasets, 
\textbf{(ii)} for the \ac{RW}-IA (CR) as well as \ac{RW}-IA (DM) datasets, both bicubic and Lanczos interpolation have accuracies and MCC up to 70.89\% and 0.42 respectively while the deep learning models have accuracies and MCC up to 73.34\% and 0.46 respectively, 
\textbf{(iii)} There is a roughly 6\% accuracy improvement for the \ac{SYN} data for the deep learning models compared to the bicubic and lanczos models and this improvement drops to about 3\% for \ac{RW} data,  
\textbf{(iv)} the performance of all the models remain consistent across both \ac{RW}-IA datasets and \textbf{(v)} in \figref{fig:RW_ROC_Curves}, we observe that there is a high degree of overlap among the \ac{ROC} curves for the data-driven models.

From (i) and (iv) we can conclude that \ac{SYN} data is more intricate than \ac{RW} data. This implies that the benefits yielded by training with \ac{SYN} dataset, while significant, is not as prominent in the \ac{RW} Iowa datasets. 
(i), (iii) and (v) implies that while \ac{SYN} data is not an exact replacement for \ac{RW} data, it provides a rather significant edge, which is all the more important when there is insufficient \ac{RW} for training. From (ii) we can conclude that the three proposed data driven models outperform classical super-resolution techniques such as bicubic and lanczos, conclusion supported by the \ac{ROC} curves in Figure \ref{fig:RW_ROC_Curves} for whom the data-driven models, in general, lie above the non-data-driven alternatives. Observation (iv) shows that  for the climatically similar \ac{RW}-Iowa(CR) and \ac{RW}-Iowa(DM) regions, training on \ac{SYN} Iowa data does indeed provide an edge. 


\subsection{Effectiveness of data-driven approaches}

In order to evaluate the effectiveness of ML models in the downscaling task, we compare performances of our candidate models to Lanczos and bicubic interpolation methods by looking at figures of some sample predictions from Iowa (Figure \ref{fig:RWIowaDesMoines}), performance comparison in the region of Iowa in Table \ref{tab:IowaResults} and the ROC curves in Figure \ref{fig:RW_ROC_Curves} for \ac{RW} data. We notice that 
\textbf{(vi)} For RW-IA (DM) samples, the deep learning models maintain a higher degree of spatial continuity in the predicted \ac{FIM}, 
\textbf{(vii)} We observe that  bicubic and Lanczos interpolation produces over-smoothed \ac{FIM} reconstructions, while the plain \ac{RDN}, \ac{RCAN} and \ac{ESRT} models are more detail-inclusive. Similar conclusions can be drawn upon inspecting the \ac{ROC} curves in Figure \ref{fig:RW_ROC_Curves} and 
\textbf{(viii)} For RW-IA (CR), the ML models show a performance improvement of 3.06\% when comparing the best ML model and non-data-driven method and, while for RW-IA (DM) there is a performance improvement of 2.45\%.

Figures \ref{fig:EUSamples} and \ref{fig:RWIowaDesMoines} show the spatial disparity among the models whose details are often obscured in aggregated metrics such as accuracy. (vi) This implies that these data-driven models are better are recognizing an underlying stream network geometry than the classical methods. However, when it comes to narrow river streams, all the models struggle capturing the nuances of the \ac{FIM} resultant from localized high elevation features such as small islands within rivers or man-made structures. (vii) shows a clear advantage of our data-driven approaches over the non-data-driven alternatives. (viii) indicates the benefits of the data-driven models when evaluated over Iowa.

\subsection{Applicability of our models to external regions}

To evaluate how transferable our models are, we draw conclusions from figures of the sample predictions from Western Europe (Figure \ref{fig:EUSamples}) and Ghana (Figure \ref{fig:GhanaSamples}) as well as the performance comparison in Table \ref{tab:ExternalResults}. We notice that 
\textbf{(ix)} for Ghana all of the models fail to adequately inundate the pixels over separated areas on account of several disconnected regions of inundation in the chosen area,
\textbf{(x)} the ML models outperform non-data driven methods for RW-EU, 
\textbf{(xi)} for the RW-EU dataset, there is an improvement of 4.89\% when comparing the accuracy of the best data- and non-data-driven methods, 
\textbf{(xii)} For RW-RR and RW-GH, there is marginal improvement (up to 0.77\% in accuracy) of the ML methods over the non-data driven methods and 
\textbf{(xiii)} For RW-EU, we notice that the ML models produce more connected streams over the non-data-driven models. 

(x) and (xi) implies that the models are transferable when considering hydroclimaticalogically similar regions since Iowa and the Meuse river in Europe lie within mid temperate zones. Similar to the observation (vi) for RW-IA (DM), (xiii) implies that the benefits of the ML model in identifying underlying network streams is also transferable to hydroclimatologically similar regions. In contrast, (xii) and (ix) both imply that the trained ML models struggle to generalize to RW-RR \& RW-GH. We speculate that this may be due to the significant differences in geography and climate when compared to Iowa. 


Our study directly implies that good quality synthetic data can be useful surrogates for downscaling low-resolution \acp{WFM} to high-resolution \acp{FIM} in regions, where such data are hard to come by, even when downscaling by a factor of 10. We noticed that such models were readily transferable to climatically similar regions as the region of training. However, Such derived ML models did not feature significantly different transferability when evaluated over hydroclimatologically dissimilar regions, which we attribute to different flood inundation characteristics, primarily at finer scales. A possible avenue to circumvent such issues is to explore additional training approaches that fall under the general area of domain adaptation. Nevertheless, data-driven models are still advantageous (and, hence, preferable) over non-data-driven alternatives in transfer scenarios like the one we considered here.

    \section{Conclusion}
In this work we study a 10$\times$ data-driven super-resolution framework for converting \acp{WFM} into high resolution \acp{FIM}. High resolution \acp{FIM} help in the study of flood inundation dynamics for a variety of applications, including real-time
emergency response. We propose three candidate ML models to produce \acp{FIM}. 
We also embellished the classical binary cross entropy loss with a soft constraint to enforce a loose satisfaction of the fractions in the \ac{WFM}. 
To circumvent data scarcity, we train our proposed ML models using HEC-RAS simulations over the region of Iowa as a stand-in for real world data. To determine the efficacy of our proposed downscaling models, we evaluate the model over five regions, three with hydroclimatical similarity to the training dataset -- Des Moines in Iowa, Cedar River in Iowa and the Meuse river in Western Europe -- and two dissimilar regions -- Red River of the north and Nasia river in Ghana. Our results indicate that, for geomorphological and hydroclimatological similar regions, a model trained on synthetic data yields benefits over traditional interpolation techniques for downscaling. This suggests that such synthetic data can act as a stand-in for training such data-intensive ML models. 
When extending these models to other regions, we notice that the benefits of synthetic data are less evident. It appears that training separate models per topographically similar regions may be the only recourse. 

Note that our study focuses solely on riverine flooding and does not address pluvial or coastal  inundation, this is left to future work. Additionally, we expect that a meaningful incorporation of topographic features to the deep learning architecture such as \acp{DEM} -- as done in \cite{Li2022VIIRSDownscaling} -- can help improve the downscaling performance. While this work was evaluated on data from Landsat, this methodology can easily be extended to other satellite products. The combination of the proposed method applied to coarse resolution optical data available daily with high-res SAR-based \acp{FIM} from Sentinel observations offers a unique opportunity for global scale flood inundation monitoring at high-resolution.

    \appendices
    \section*{Acknowledgment}

The authors acknowledge partial support from an Institutional Research Incentive seed grant titled ``\textit{Satellite Imagery Downscaling via Deep Image Super-resolution}'' and provided by the College of Engineering \& Sciences of Florida Institute of Technology. Nikolopoulos acknowledges support from NOAA grant (NA20OAR4600288) titled ``\textit{Accelerate the Exploitation of Satellite Observations to Improve Flooding and Inundation Monitoring and Forecasts}''. We would like to acknowledge high-performance computing support from Casper (doi:10.5065/D6RX99HX) provided by NCAR’s Computational and Information Systems Laboratory, sponsored by the National Science Foundation. The authors acknowledge Dr. Emanuel Storey for creating the Landsat based \acp{FIM} for the flood events in Red River, Iowa and Europe. The author also acknowledge the Iowa Flood Center for providing the outputs from the physics-based simulations. 
    \ifCLASSOPTIONcaptionsoff
      \newpage
    \fi
    \bibliographystyle{IEEEtran}
    \bibliography{references.bib}
%

\begin{IEEEbiographynophoto}{Akshay Aravamudan}
is a PhD student at the Florida Institute of Technology. His research areas of interest include machine learning, stochastic point processes for the study of information diffusion and influence characterization in social media, machine learning for hydrology and machine learning on the edge. Before joining FIT’s Center for Advanced Data Analytics \& Systems (CADAS), he obtained his M.S. in Computer Engineering at the Florida Institute of Technology.
\end{IEEEbiographynophoto}

\begin{IEEEbiographynophoto}{Zimeena Rasheed}
is a PhD student at Rutgers University, New Brunswick. She graduated with a M.S. in Civil Engineering from Florida Institute of Technology in 2020. Her research broadly includes the study of stream flows, their prediction for anticipation of floods as well as satellite-based flood inundation. 
\end{IEEEbiographynophoto}

\begin{IEEEbiographynophoto}{Xi Zhang}
is a PhD student at the Florida Institute of Technology. Her research interests include probablistic modeling, statistical methods, ML/AI with applications in social media, cyber-security and the earth sciences. She received her M.S. in Electrical Engineering at Florida Institute of Technology, and she is currently a member of FIT’s Center for Advanced Data Analytics \& Systems (CADAS).
\end{IEEEbiographynophoto}

\begin{IEEEbiographynophoto}{Kira E. Scarpignato}
is expected to graduate with a B.S in Biomedical Engineering and a minor in Biology in May 2024 from Florida Institute of Technology. Her current research is focused on cardiovascular tissue engineering.
\end{IEEEbiographynophoto}

\begin{IEEEbiographynophoto}{Efthymios I. Nikolopoulos}
is an Associate Professor of Civil \& Environmental Engineering at Rutgers University, New Brunswick. He received his Engineering Diploma from the Technical University of Crete, Greece, his M.Sc. degree from the University of Iowa, and his Ph.D. degree from the University of Connecticut, U.S. He has also worked as a postdoc at the University of Padova, Italy. His expertise is in the modeling and monitoring of hydrometeorological and hydrologic extremes (extreme precipitation, floods, droughts, debris flows). His main research goal is to improve the understanding and predictability of hydrologic extremes and develop methods to mitigate their impacts. 
\end{IEEEbiographynophoto}

\begin{IEEEbiographynophoto}{Witold F. Krajewski}
received his Ph.D. from Warsaw University of Technology in 1980. He is the Rose \& Joseph Summers Chair in Water Resources Engineering and Professor in Civil and Environmental Engineering at the University of Iowa. He serves as Director of the Iowa Flood Center, an entity funded by the State of Iowa in the United States and housed at the University of Iowa. His research interests include all aspects of flood forecasting. He is a Fellow of the American Meteorological Society (AMS), the American Geophysical Union (AGU), and a member of the U.S. National Academy of Engineering.

\end{IEEEbiographynophoto}
\begin{IEEEbiographynophoto}{Georgios C. Anagnostopoulos}
is an Associate Professor of Electrical \& Computer Engineering at the Florida Institute of Technology in Melbourne, Florida. He received is Engineering Diploma from the University of Patras in 1994 and his M.Sc. and Ph.D. degrees in Electrical Engineering from the University of Central Florida in 1997 and 2001 respectively. His areas of expertise are machine learning, modeling and optimization. He is a senior member of the IEEE.
\end{IEEEbiographynophoto}




\end{document}